\pdfoutput=1

\documentclass[11pt]{article}

\usepackage[final]{acl}

\usepackage{times}
\usepackage{latexsym}

\usepackage[T1]{fontenc}

\usepackage[utf8]{inputenc}

\usepackage{microtype}

\usepackage{inconsolata}

\usepackage{graphicx}

\usepackage{booktabs}
\usepackage{multirow}
\usepackage{wrapfig}
\usepackage{amsmath}

\usepackage{xcolor}
\usepackage{colortbl}
\definecolor{darkgreen}{RGB}{0,100,0}
\definecolor{lightgray}{gray}{0.9}
\usepackage{array} 

\usepackage{booktabs,tabularx,makecell}
\newcolumntype{Y}{>{\centering\arraybackslash}X} 

\newcolumntype{C}[1]{>{\centering\arraybackslash}p{#1}}

\usepackage[most]{tcolorbox}
\usepackage{enumitem}
\tcbset{
  boxrule=0.3pt, colframe=black!35, colback=black!2,
  left=6pt, right=6pt, top=4pt, bottom=4pt, arc=2pt
}

\usepackage{float}
\usepackage{subcaption}
\usepackage{afterpage}

\newcommand\blfootnote[1]{%
  \begingroup
  \renewcommand\thefootnote{}\footnote{#1}%
  \addtocounter{footnote}{-1}%
  \endgroup
}

\title{Benchmarking Direct Preference Optimization for\\Medical Large Vision--Language Models}

\author{
    Dain Kim$^{1,*}$ \quad Jiwoo Lee$^{1,*}$ \quad Jaehoon Yun$^{2,3}$ \quad Yong Hoe Koo$^{4}$\\ \textbf{Qingyu Chen}$^{5}$ \quad \textbf{Hyunjae Kim}$^{5,\dagger}$ \quad \textbf{Jaewoo Kang}$^{1,2,\dagger}$ \\
    $^1$Korea University \quad 
    $^2$AIGEN Sciences \quad $^3$Hanyang University College of Medicine \\
    $^4$Asan Medical Center, University of Ulsan College of Medicine \quad
    $^5$Yale University
    \\
    \texttt{\{dain-kim,hijiwoo7,kangj\}@korea.ac.kr}\quad\texttt{hyunjae.kim@yale.edu}\\
}

\begin{document}
\maketitle
\begin{abstract}

Large Vision-Language Models (LVLMs) hold significant promise for medical applications, yet their deployment is often constrained by insufficient alignment and reliability. While Direct Preference Optimization (DPO) has emerged as a potent framework for refining model responses, its efficacy in high-stakes medical contexts remains underexplored, lacking the rigorous empirical groundwork necessary to guide future methodological advances.
To bridge this gap, we present the first comprehensive examination of diverse DPO variants within the medical domain, evaluating nine distinct formulations across two medical LVLMs: LLaVA-Med and HuatuoGPT-Vision. 
Our results reveal several critical limitations: current DPO approaches often yield inconsistent gains over supervised fine-tuning, with their efficacy varying significantly across different tasks and backbones. 
Furthermore, they frequently fail to resolve fundamental visual misinterpretation errors. 
Building on these insights, we present a targeted preference construction strategy as a proof-of-concept that explicitly addresses visual misinterpretation errors frequently observed in existing DPO models. This design yields a 3.6\% improvement over the strongest existing DPO baseline on visual question-answering tasks.
To support future research, we release our complete framework, including all training data, model checkpoints, and our codebase at \url{https://github.com/dmis-lab/med-vlm-dpo}.

\end{abstract}

\newcommand{\draftcomment}[3]{%
  \textcolor{#2}{\textbf{[#1: #3]}}%
}

\newcommand{\llavamed}{LLaVA-Med}
\newcommand{\huatuo}{HuatuoGPT-Vision}

\newcommand{\todo}[1]{\draftcomment{TODO}{red}{#1}}

\newcommand{\hyunjae}[1]{\draftcomment{hyunjae}{teal}{#1}}
\newcommand{\dain}[1]{\draftcomment{dain}{brown}{#1}}
\newcommand{\jiwoo}[1]{\draftcomment{jiwoo}{purple}{#1}}

\blfootnote{\textsuperscript{*}These authors contributed equally to this work.}
\blfootnote{\textsuperscript{$\dagger$}Corresponding authors.}



\section{Introduction}
\label{intro}

Recent advances in Large Vision-Language Models (LVLMs), which integrate powerful large language models (LLMs) with visual encoders, have greatly improved AI's ability to process and reason over multimodal inputs~\cite{alayrac2022flamingo, li2023blip, liu2023visual, zhu2023minigpt, openai2023gpt4v}.
In the medical domain, these advances have enabled applications such as diagnostic support, clinical question answering, and report generation~\cite{kline2022multimodal, li2023llava, chen2024huatuogpt,wu2023towards, xie2024medtrinity}, but safe deployment remains a critical challenge.
For instance, factually incorrect or fabricated outputs, often described as hallucinations, pose particular risks~\cite{maynez2020faithfulness, liu2024survey, kim2025medical}.
Additionally, errors in interpreting medical images~\cite{jin2024hidden} may lead to cascading failures in downstream decision-making~\cite{zhang2024language}.

Direct Preference Optimization (DPO)~\cite{rafailov2023direct} and its subsequent variants have been explored to improve the reliability of language models~\cite{ethayarajh2024kto,xu2024contrastive}.
By leveraging preference signals to contrast output pairs, DPO optimizes model parameters to favor safer and more faithful generations.
However, while these approaches have been predominantly validated in general-domain language and vision-language tasks~\cite{saeidi2025insights,zhou2024aligning,wang2024mdpo}, their performance in high-stakes fields such as medicine remains insufficiently understood.
Given the distinct data characteristics and the specific nature of medical errors, general-domain optimizations may not directly translate to reliable clinical performance, necessitating a dedicated validation of preference-based alignment within this specialized context.

\begin{figure*}[t!]
    \centering
    \includegraphics[width=1.0\linewidth]{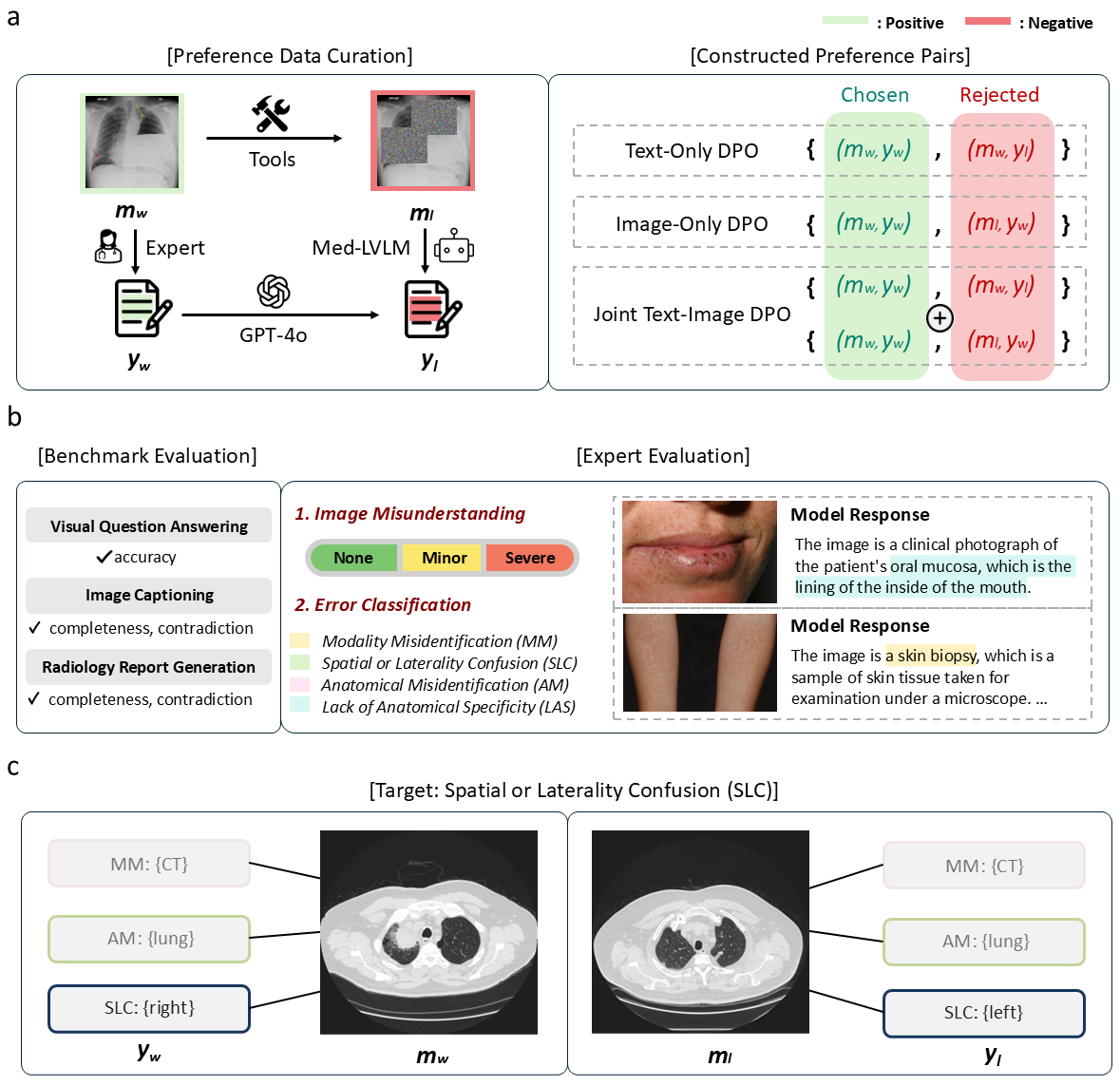} 
    \caption{{Overview of the study design.} 
    \textbf{a.} Evaluated DPO models: Illustration of three DPO configurations (text-only, image-only, and joint text-image) categorized by the modality contrasted during preference learning. 
    \textbf{b.} Evaluation framework: A multi-faceted assessment combining automated benchmark evaluation across three core tasks (visual question answering, image captioning, and radiology report generation) and expert qualitative analysis focused on image misunderstanding severity and specific error types (e.g., MM, SLC, AM, LAS).
    \textbf{c.} Targeted preference pair construction: A demonstration of our approach using a SLC example. We construct contrastive pairs by perturbing text keywords and retrieving corresponding ``hard-negative'' images to improve the model's spatial and anatomical grounding (see Section~\ref{sec:error_specific} for details).}
    \label{fig:main_fig}
\end{figure*}

In this paper, we present the first comprehensive evaluation of DPO-based alignment for medical LVLMs. 
We systematically analyze leading multi-modal DPO methods from both general and medical domains, categorizing them into three distinct groups based on their data perturbation strategies: text-only, image-only, and joint text-image (Figure~\ref{fig:main_fig}a). 
We implement nine DPO formulations atop two representative medical LVLMs: \llavamed~\cite{li2023llava} and \huatuo~\cite{chen2024huatuogpt}. 

We analyze the models in two stages: a benchmark evaluation and an expert evaluation (Figure~\ref{fig:main_fig}b). For benchmark evaluation, we first compile five datasets spanning both visual question answering (VQA) and two generation tasks: radiology report generation and image captioning. 
We evaluate baseline models using accuracy for VQA and employ an LLM-as-a-judge framework~\cite{zheng2023judging,gu2024survey} to assess completeness and contradiction in the generation tasks.
We observe that all DPO variants consistently improve VQA accuracy. However, similar gains can also be achieved through standard supervised fine-tuning (SFT), making the advantage of DPO less evident in this setting. In the generation tasks, no single method consistently outperforms others across tasks or metrics. For example, a text-only DPO model achieved the highest completeness score on the image captioning dataset with a 3.11\% improvement, yet exhibited a 4.81\% decrease in report generation performance.
Overall, the results suggest that the effects of DPO may not fully align with previous reports of its effectiveness in general-domain~\cite{zhou2024aligning,wang2024mdpo} or early medical-domain studies~\cite{zhu2024mmedpo}. 

To gain deeper insights, we conduct a manual error analysis, aiming to uncover the qualitative limitations underlying our quantitative findings.
We observe that a substantial majority of errors originated from the misinterpretation of medical images. 
Notably, the base LLaVA-Med model exhibits image misunderstandings in 90\% of its image captioning outputs (82.5\% severe, 7.5\% minor) and 97.5\% for its report generation outputs (82.5\% severe, 15\% minor).
While a DPO model significantly mitigates the most critical failures---reducing severe interpretation errors from 90\% to 50\% in image captioning---this improvement is accompanied by a marked increase in minor misinterpretations, which rises from 7.5\% to approximately 30\%. 
While encouraging, these results suggest that current DPO formulations merely shift the error profile from severe to minor rather than fully resolving the underlying issues.

To investigate whether these gaps could be narrowed through more targeted alignment, we identify four major categories of visual misinterpretation errors recurring in model outputs. 
We then tailor the DPO training process by constructing preference data specifically designed to counteract these errors, exploring the feasibility of domain-targeted preference modeling (Figure~\ref{fig:main_fig}c).
While previous experiments showed DPO providing only marginal improvements over SFT in VQA tasks, our specialized DPO approach yields consistent performance gains, outperforming the base LLaVA-Med model by 6.9\%, the SFT model by 4.6\%, and the best-performing baseline DPO model by 3.6\%.

Beyond our initial findings, the observed limitations underscore the need for more rigorous community-wide validation and the development of robust, domain-aware alignment strategies for medical AI. 
To support these efforts and foster further innovation, we publicly release our entire framework, including the code, curated datasets, trained models, and expert-verified error annotations.\footnote{\url{https://github.com/dmis-lab/med-vlm-dpo}} 

\section{Related Work}
\label{sec:background}

\subsection{Large Vision-Language Models in Medicine}
Medical LVLMs are adapted from general-purpose models through fine-tuning on biomedical data~\cite{li2023llava, chen2024huatuogpt, zhang2024generalist, lin2025healthgpt,sellergren2025medgemma}, typically using image-caption pairs from PubMed Central for visual alignment and human- or LLM-generated prompts for instruction tuning. 
Some models further integrate biomedical-specific vision encoders~\cite{lin2023pmc, zhang2025multimodal} to enhance domain relevance. While models like LLaVA-Med~\cite{li2023llava} and HuatuoGPT-Vision~\cite{chen2024huatuogpt} support broad medical tasks, others are tailored to specific fields such as radiology~\cite{wu2023towards, chen2024chexagent}, surgery~\cite{wang2024surgical}, pathology~\cite{seyfioglu2024quilt}, and dermatology~\cite{zhou2024pre}.

\subsection{Direct Preference Optimization for LVLMs}
\label{subsec:dpo_lvml}

Direct Preference Optimization (DPO) was originally proposed for text-only preference pairs~\cite{rafailov2023direct}. 
Subsequent works have extended this framework to multimodal tasks by modifying the definition of the contrastive objective.
Specifically, variants differ in the modality being contrasted: some apply DPO to output text only, while others contrast both input and output jointly. 
HA-DPO~\cite{zhao2023beyond} and HSA-DPO~\cite{xiao2025detecting} generate rejections by automatically detecting and correcting hallucinated spans, while Silkie~\cite{li2023silkie} ranks multiple model outputs to form preference pairs. 
SIMA~\cite{wang2025enhancing} further leverages self-feedback, where the model compares and critiques its own outputs, while CLIP-DPO~\cite{ouali2024clipdpo} derives preference signals from image-text similarity scores provided by a pretrained CLIP model.
In contrast, several methods contrast both outputs and inputs simultaneously. 
mDPO~\cite{wang2024mdpo} integrates text-based rejections with corrupted inputs, using random cropping as the perturbation. 
POVID~\cite{zhou2024aligning} combines GPT-4-generated hallucinations on the text side with Gaussian-noised images on the visual side. 
MMedPO~\cite{zhu2024mmedpo} likewise merges modalities, treating hallucinated responses as rejections while contrasting original images against ROI-noised counterparts.
                                                                        
\begin{table*}[ht!]
\centering
\footnotesize
\begin{tabular}{lll}
\toprule
\textbf{Method} & \textbf{Description} & \textbf{Relevant Methods} \\
\midrule
\multicolumn{3}{l}{\textit{\textbf{Text-only Perturbation}}} \\
\midrule
Text-Hallu & $y_w$ corresponds to $y$. & \multirow{4}{*}{\begin{tabular}[c]{@{}l@{}}POVID~\cite{zhou2024aligning},\\ MMedPO~\cite{zhu2024mmedpo}\end{tabular}} \\
& $y_l$ is generated by hallucinating $y$ by GPT-4o. &  \\
\cmidrule(lr){2-2}
Text-Hallu + NLL & Text-Hallu with the addition of NLL loss. & \\
\midrule
Text-Noise & $y_w$ corresponds to $y$. & \multirow{4}{*}{\begin{tabular}[c]{@{}l@{}}POVID~\cite{zhou2024aligning},\\ STIC~\cite{deng2024enhancing}\end{tabular}} \\
& $y_l$ is self-generated from the image $m$ with Gaussian noise. &
 \\
\cmidrule(lr){2-2}
Text-Noise + NLL & Text-Noise with the addition of NLL loss. & \\
\midrule
IRPO & $y_w$ is a self-generated response closely aligned with $y$. & IRPO~\cite{pang2024iterative} \\
& $y_l$ is a less aligned one. \\
\midrule
\multicolumn{3}{l}{\textit{\textbf{Image-only Perturbation}}} \\
\midrule
Image-Noise & $m_w$ corresponds to $m$. & \multirow{4}{*}{\begin{tabular}[c]{@{}l@{}}mDPO~\cite{wang2024mdpo},\\ POVID~\cite{zhou2024aligning},\\
MMedPO~\cite{zhu2024mmedpo}\end{tabular}}
 \\
           & $m_l$ corresponds to $m$ with Gaussian noise. &  \\
\cmidrule(lr){2-2}
Image-ROI & $m_w$ corresponds to $m$. &  \\
            & $m_l$ corresponds to $m$ with Gaussian noise applied to ROI. & \\
\midrule
\multicolumn{3}{l}{\textit{\textbf{Joint Text-Image Perturbation}}} \\
\midrule
mDPO & $y_w$ and $m_w$ correspond to $y$ and $m$, respectively. & mDPO~\cite{wang2024mdpo} \\
     & $m_l$ corresponds to $m$ with random cropping applied. \\
     & $y_l$ is self-generated from $m_l$.\\
\midrule
MMedPO & $y_w$ corresponds to $y$, and $y_l$ is generated by GPT-4o. & MMedPO~\cite{zhu2024mmedpo} \\
     & $m_w$ and $m_l$: Same as Image-ROI. & \\
\bottomrule
\end{tabular}
\caption{Categorization of DPO methods. $y$: ground-truth response.
$m$: original image.
$y_w$ and $y_l$: preferred (chosen) and dispreferred (rejected) responses, respectively.
$m_w$ and $m_l$: preferred (chosen) and dispreferred (rejected) images.
Relevant methods denote existing frameworks whose core principles were adapted and tuned for medical multimodal alignment.
ROI: Regions of Interest.
}
\label{tab:dpo_methods}
\end{table*}

\subsection{Benchmarking Medical LVLMs}

A growing body of work has sought to benchmark medical LVLMs across various tasks and dimensions. MultiMedEval~\cite{royer2024multimedeval} and Asclepius~\cite{liu2024spectrum} offer large-scale suites to evaluate accuracy across modalities and specialties, addressing prior issues of fragmented evaluation practices. In parallel, CARES~\cite{xia2024cares} introduces a multidimensional framework for trustworthiness, covering trustfulness, fairness, safety, privacy, and robustness. MedHEval~\cite{chang2025medheval} and Med-HallMark~\cite{chen2024medhallmark} further systematize hallucination evaluation and highlight domain-specific risks like visual misinterpretation and knowledge deficiency.
Yet, most benchmarks focus on off-the-shelf models; only MedHEval~\cite{chang2025medheval} systematically examines inference-time hallucination mitigation.
To the best of our knowledge, no prior work has comprehensively evaluated DPO models for medical LVLMs. Our work examines whether and how such preference-based tuning methods affect medical LVLM behavior, through both automatic benchmarks and structured expert assessments.







\section{Evaluated DPO Models}

Prior work on DPO for LVLMs can be categorized based on the modality being contrasted: text-only, image-only, or joint text-image, as illustrated in Figure~\ref{fig:main_fig}a.
Building on this perspective, we further organize the landscape along two axes: (i) the underlying training objective and (ii) the preference pair curation strategy.

Most of existing approaches were originally developed for the general domain and mainly address issues such as object hallucination~\cite{bai2024hallucination}, which are not directly applicable to medical data. To bridge this gap, we adapted these methods to the medical domain while retaining their core principles, resulting in eight domain-specific DPO variants.
We also include MMedPO~\cite{zhu2024mmedpo}, a method developed specifically for medical applications, yielding a total of nine models.
Please refer to Table~\ref{tab:dpo_methods} for the full list.
In Appendix~\ref{sec:appendix}, we provide illustrative examples of preference pairs.

\subsection{Text-only Perturbation}
Let $q$ be a text prompt (i.e., the instruction or query to the model), $y_w$ the preferred (chosen) response, and $y_l$ the dispreferred (rejected) response.
The standard DPO objective is:
\begin{equation}
\begin{aligned}
\mathcal{L}_{\mathrm{DPO}} = & - \log \sigma \Bigg( \beta \log \frac{\pi_\theta(y_w \mid q)}{\pi_{\mathrm{ref}}(y_w \mid q)} \\
& \quad - \beta \log \frac{\pi_\theta(y_l \mid q)}{\pi_{\mathrm{ref}}(y_l \mid q)} \Bigg),
\end{aligned}
\label{eq:dpo}
\end{equation}
where $\pi_\theta$ is the target policy, $\pi_{\mathrm{ref}}$ is a fixed reference model (typically a supervised fine-tuned checkpoint), $\beta$ is a temperature-like scaling factor, and $\sigma(\cdot)$ denotes the sigmoid function.

In multimodal settings with an input image $m$ and prompt $q$, this extends to:
\begin{equation}
\begin{aligned}
\mathcal{L}_{\mathrm{DPO}_m} = & - \log \sigma \Bigg( 
\beta \log \frac{\pi_\theta(y_w \mid m, q)}{\pi_{\mathrm{ref}}(y_w \mid m, q)} \\
& \quad - \beta \log \frac{\pi_\theta(y_l \mid m, q)}{\pi_{\mathrm{ref}}(y_l \mid m, q)} 
\Bigg)
\end{aligned}
\label{eq:multimodal_dpo}
\end{equation}

The text-only models differ in how $y_w$ and $y_l$ are defined. 
In Text-Hallu, $y_w$ is the ground-truth response $y$, while $y_l$ is generated by GPT-4o with induced hallucinations. 
In Text-Noise, $y_w=y$, and $y_l$ is self-generated from a Gaussian-noised image $m$. 
In IRPO, $y_w$ and $y_l$ pairs are selected from $N{=}20$ self-generated responses (temperature 1.2) ranked by ROUGE-L against the ground truth, with the top-1 and bottom-1 responses chosen.
The ``+NLL'' variants of Text-Hallu and Text-Noise further incorporate a negative log-likelihood term to encourage higher probability for $y$.
The IRPO objective inherently includes an NLL term within its formulation (see Appendix~\ref{sec:appendix} for details).

\subsection{Image-only Perturbation}

In this setting, the image $m$ is perturbed, whereas the prompt $q$ and reference response $y$ remain identical across conditions. 
That is, both $(m_w, q)$ and $(m_l, q)$ are paired with the same $y$, ensuring that differences arise solely from the image modality. 
The Conditional Preference Optimization (CoPO) loss~\cite{wang2024mdpo} applies a contrastive objective over image perturbations:
\begin{equation}
\begin{aligned}
\mathcal{L}_{\mathrm{CoPO}} = & - \log \sigma \Bigg( 
\beta \log \frac{\pi_\theta(y_w \mid m_w, q)}{\pi_{\mathrm{ref}}(y_w \mid m_w, q)} \\
& \quad - \beta \log \frac{\pi_\theta(y_w \mid m_l, q)}{\pi_{\mathrm{ref}}(y_w \mid m_l, q)} 
\Bigg).
\end{aligned}
\label{eq:copo}
\end{equation}
Here, Image-Noise perturbs $m$ with Gaussian noise, while Image-ROI perturbs the ROI (Regions of Interest) extracted using MedCLIP~\cite{wu2023medklip}.

\subsection{Joint Text-Image Perturbation}
Models in this group combine the objectives defined in Equations~\ref{eq:multimodal_dpo} and~\ref{eq:copo}. 
In mDPO, $y_w$ and $m_w$ are the ground-truth response $y$ and original image $m$, while $m_l$ is a randomly cropped version of $m$. 
The rejected response $y_l$ is conditioned on the corrupted image $m_l$.
In MMedPO, $y_w$ and $m_w$ are the ground-truth response $y$ and original image $m$.
The rejected response $y_l$ is generated by GPT-4o.
The rejected image $m_l$ are defined as in Image-ROI, with Gaussian noise applied to the ROI. 
Please refer to Appendix~\ref{sec:appendix} for the detailed mathematical formulations for mDPO and MMedPO.

\section{Experiments}
\begin{table}[t]
  \centering
  \footnotesize
    \begin{tabular}{llll}
      \toprule
      \textbf{Dataset} & \textbf{\# Ex.} & \textbf{\# Img.} & \begin{tabular}[c]{@{}l@{}}\textbf{Modality}\end{tabular}  \\
      \midrule
      \multicolumn{4}{l}{\textit{\textbf{Visual Question Answering (VQA)}}} \\
      {VQA-RAD} & 451 & 204 & Radiology  \\
      {SLAKE} & 1,061 & 96 & Radiology \\
      {PathVQA} & 6,719 & 858 & Pathology \\
      \midrule
      \multicolumn{4}{l}{\textit{\textbf{Image Captioning}}} \\
      {AMBOSS} & 164 & 164 & Misc. \\
    \midrule
      \multicolumn{4}{l}{\textit{\textbf{Radiology Report Generation}}} \\
      {MIMIC-CXR} & 1,031 & 1,031 & Chest X-ray \\
      \bottomrule
    \end{tabular}
  \caption{
    Datasets used in the benchmark evaluation.
    \# Ex. and  \# Img.: the number of examples and images, respectively.
  }
  \label{tab:benchmark}
\end{table}

\begin{table*}[t]
\centering
\footnotesize
\setlength{\tabcolsep}{6pt}
\renewcommand{\arraystretch}{1.05}
\begin{tabular}{>{\raggedright\arraybackslash}p{2.4cm} l ll ll}
\toprule
 & \multicolumn{1}{l}{\textbf{VQA}}
 & \multicolumn{2}{l}{\textbf{Image Captioning}}
 & \multicolumn{2}{l}{\textbf{Radiology Report Generation}} \\
\cmidrule(lr){2-2}\cmidrule(lr){3-4}\cmidrule(lr){5-6}
\textbf{Model} & \textbf{Acc (↑)} & \textbf{Comp (↑)} & \textbf{Cont (↓)} & \textbf{Comp (↑)} & \textbf{Cont (↓)} \\
\midrule
\multicolumn{6}{l}{\textbf{\textit{LLaVA-Med}}} \\
\midrule
Base Model                      & 38.7 (-) & 9.11 (-)  & 19.56 (-) & 15.83 (-) & 11.90 (-) \\
\midrule
SFT                       & 41.0 (+2.3) & 9.66 (+0.55) & 20.43 (+0.87) & 10.96 (-4.87) & 13.64 (+1.74) \\
\midrule
Text-Hallu               & 41.3 (+2.6) & \textbf{12.22 (+3.11)} & 13.93 (-5.63) & 11.02 (-4.81) & 12.39 (+0.49) \\
\hspace*{0.9em}+ NLL      & \textbf{42.0 (+3.3)} & 10.67 (+1.56) & 19.15 (-0.41) & 13.12 (-2.71) & \textbf{11.03 (-0.87)} \\
Text-Noise                & 39.0 (+0.3) & 9.80 (+0.69) & 11.63 (-7.93) & 17.70 (+1.87) & 11.99 (+0.09) \\
\hspace*{0.9em}+ NLL      & 40.6 (+1.9) & 9.85 (+0.74) & 21.23 (+1.67) & 13.13 (-2.70) & 13.63 (+1.73) \\
IRPO                      & 39.0 (+0.3) & 9.12 (+0.01) & 19.79 (+0.23) & \textbf{18.33 (+2.50)} & 12.00 (+0.10) \\
\midrule
Image-Noise                & 40.0 (+1.3) & 8.22 (-0.89) & \textbf{9.35 (-10.21)} & 12.24 (-3.59) & 12.77 (+0.87) \\
Image-ROI               & 41.0 (+2.3) & 10.74 (+1.63) & 11.38 (-8.18) & 11.34 (-4.49) & 13.23 (+1.33) \\
\midrule
mDPO                      & 41.9 (+3.2) & 10.44 (+1.33) & 20.58 (+1.02) & 12.75 (-3.08) & 11.98 (+0.08) \\
MMedPO                    & 40.1 (+1.4) & 11.38 (+2.27) & 12.75 (-6.81) & 10.82 (-5.01) & 12.74 (+0.84) \\
\midrule
\multicolumn{6}{l}{\textbf{\textit{HuatuoGPT-Vision}}} \\
\midrule
Base Model                    & 49.1 (-) & 20.97 (-) & 29.58 (-) & 21.81 (-) & 23.05 (-) \\
\midrule
SFT                       & 51.9 (+2.8) & 21.03 (+0.06) & 27.15 (-2.43) & 21.91 (+0.10) & 22.76 (-0.29) \\
\midrule
Text-Hallu               & \textbf{53.2 (+4.1)} & 21.48 (+0.51) & 26.86 (-2.72) & 22.34 (+0.53) & 22.93 (-0.12) \\
\hspace*{0.9em}+ NLL      & 51.7 (+2.6) & 20.39 (-0.58) & 28.81 (-0.77) & 22.36 (+0.55) & 22.60 (-0.45) \\
Text-Noise                & 51.1 (+2.0) & 20.78 (-0.19) & 25.88 (-3.70) & 22.50 (+0.69) & 25.14 (+2.09) \\
\hspace*{0.9em}+ NLL      & 52.4 (+3.3) & 20.89 (-0.08) & 25.00 (-4.58) & 22.05 (+0.24) & 24.15 (+1.10) \\
IRPO                      & 52.7 (+3.6) & 20.97 (+0.00) & 26.83 (-2.75) & 22.10 (+0.29) & \textbf{21.57 (-1.48)} \\
\midrule
Image-Noise                & 49.6 (+0.5) & 22.26 (+1.29) & \textbf{24.69 (-4.89)} & \textbf{22.80 (+0.99)} & 22.98 (-0.07) \\
Image-ROI               & 52.4 (+3.3) & \textbf{23.03 (+2.06)} & 28.54 (-1.04) & 22.21 (+0.40) & 22.81 (-0.24) \\
\midrule
mDPO                      & 50.9 (+1.8) & 23.02 (+2.05) & 25.22 (-4.36) & 21.51 (-0.30) & 22.42 (-0.63) \\
MMedPO                    & 51.8 (+2.7) & 21.51 (+0.54) & 26.90 (-2.68) & 22.07 (+0.26) & 22.79 (-0.26) \\
\bottomrule
\end{tabular}
\caption{Performance comparison of DPO methods and SFT baselines across LLaVA-Med and HuatuoGPT-Vision backbones. Evaluation spans visual question answering (VQA; averaged across the SLAKE, VQA-RAD, and PathVQA datasets), image captioning (the AMBOSS dataset), and radiology report generation (the MIMIC-CXR dataset). We report accuracy (Acc) for VQA, along with completeness (Comp) and contradiction (Cont) for generation tasks (↑: higher is better; ↓: lower is better). Values in parentheses denote the performance delta relative to each SFT base model. For reference, GPT-4.1 achieves a VQA accuracy of 58.1\%.}
\label{tab:score_benchmark}
\end{table*}

\subsection{Tasks and Datasets}
\label{subsec:benchmark_construction}

The benchmark evaluation includes the following tasks and datasets, along with a description of the metrics used. See Table~\ref{tab:benchmark} for a summary.

\paragraph{Visual question answering (VQA)} This task is widely used for evaluating medical LVLMs.
For datasets, we use VQA-RAD~\cite{lau2018dataset}, SLAKE~\cite{liu2021slake}, and PathVQA~\cite{he2020pathvqa}, all of which pair radiological/pathological images with clinician-annotated QA pairs.
We report accuracy averaged across the three datasets.

\paragraph{Image captioning}
This task focuses on describing the core visual findings in medical images across various modalities, including radiology, pathology, dermatology, and endoscopy. We utilize image-caption pairs curated by medical experts, sourced from AMBOSS, a medical question bank platform designed for licensing/board exam preparation.\footnote{\url{https://www.amboss.com/us}}
While the data is licensed, we obtained explicit permission for their use in this study.
For metrics, we apply a statement-level, LLM-based evaluation: reference reports are decomposed into atomic clinical statements~\cite{min2023factscore}, and model outputs are classified into entailment, partial entailment, contradiction, or neutral (see Appendix~\ref{appendix:nli} for details). 
From this, we compute completeness (proportion of entailed statements) and contradiction scores (proportion of contradicted statements).

\paragraph{Radiology report generation}
We evaluate models on the MIMIC-CXR dataset~\cite{johnson2019mimic} using a filtered test set to ensure precise, image-grounded assessment (see Appendix~\ref{appendix:mimic_curation} for details). 
The same metrics as in image captioning are used---completeness and contradiction.

\begin{table*}[ht!]
\centering
\footnotesize
\setlength{\tabcolsep}{5pt}
\begin{tabular}{>{\raggedright\arraybackslash}p{2cm} *{3}{C{1.8cm}} *{4}{C{1.1cm}}}
\toprule
& \multicolumn{3}{c}{\textbf{Image Misunderstanding (\%)}} 
& \multicolumn{4}{c}{\textbf{Error-type Distribution (\#)}} \\
\cmidrule(lr){2-4} \cmidrule(lr){5-8}
\textbf{Model} & \textbf{None} & \textbf{Minor} & \textbf{Severe} 
& \textbf{MM} & \textbf{SLC} & \textbf{AM} & \textbf{LAS} \\
\midrule
\multicolumn{8}{l}{\textbf{\textit{Image Captioning}}} \\
\midrule
Base Model   & 10.0 & 7.5  & 82.5 & 11 & 5 & 6  & 0 \\
TxtPert-LLM    & 20.0 & 30.0 & 50.0 &  5 & 4 & 11 & 1 \\
MMedPO & 15.0 & 17.5 & 67.5 &  9 & 4 & 11 & 2 \\
SFT    & 0.0  & 15.0 & 85.0 & 11 & 2 & 12 & 1 \\
\midrule
\multicolumn{8}{l}{\textbf{\textit{Radiology Report Generation}}} \\
\midrule
Base Model   & 2.5  & 15.0 & 82.5 & 0 & 3 & 0  & 9 \\
TxtPert-LLM    & 5.0  & 35.0 & 57.5 & 0 & 1 & 0  & 22 \\
MMedPO & 2.5  & 35.0 & 60.0 & 0 & 1 & 0  & 14 \\
SFT    & 5.0  & 17.5 & 77.5 & 0 & 1 & 0  & 24 \\
\bottomrule
\end{tabular}
\caption{Expert evaluation on the image captioning and report generation tasks with {LLaVA-Med}.
Image misunderstanding is reported as percentages (\%), and error-type distribution as counts
(out of 40 evaluated cases per dataset).
MM: Modality misidentification; SLC: Spatial or laterality confusion;
AM: Anatomical misidentification; LAS: Lack of anatomical specificity.
Detailed descriptions are provided in the main text.
}
\label{tab:expert_eval}
\end{table*}

\subsection{Base LVLMs}
We use LLaVA-Med v1.5 with a Mistral-7B backbone \cite{li2023llava} and HuatuoGPT-Vision with a Qwen2-7B backbone~\cite{chen2024huatuogpt} as our base models, 
both of which were pretrained and instruction-tuned on large-scale medical data. 
We sample 10,000 instructions from each model's training corpus, which serve as a shared foundation for subsequent SFT and DPO preference pair construction.
This follows established practice in recent studies, which typically employ between 5,000 and 17,000 preference pairs \cite{wang2024mdpo, wang2025enhancing, zhou2024aligning, deng2024enhancing}.

\subsection{Benchmark Results}
We conducted bootstrapping with 100 resampling iterations of equal size.
Table~\ref{tab:score_benchmark} summarizes model performance across VQA, image captioning, and report generation tasks.
Crucially, our results reveal a significant limitation of applying existing DPO strategies to the medical domain. 
While DPO models improved VQA accuracy (+0.3--3.3\% for LLaVA-Med, +0.5--4.1\% for HuatuoGPT-Vision), comparable gains were observed with matched SFT baselines. This indicates that the perceived benefits may stem largely from additional training rather than preference optimization itself.

For generation tasks, performance was inconsistent; while some DPO variants improved specific metrics, others showed regression, with deltas often falling within the range of run-to-run variability. Contrary to prior findings reporting uniform improvements~\cite{zhu2024mmedpo}, our analysis demonstrates that naively transferring DPO methods to complex medical tasks yields unstable and marginal gains.

\subsection{Expert Evaluation}
Beyond quantitative metrics, we conducted an expert evaluation to investigate the underlying failure modes that the DPO models exhibit in medical tasks.
We specifically focused our error analysis on image misunderstanding, as accurate visual recognition serves as the critical foundation for all downstream medical interpretation. 
Since visual errors often lead to a cascade of incorrect clinical reasoning, two experts categorized these failures by their severity (e.g., none, moderate, or severe) to measure their actual impact.
Here, severe errors refer to critical misinterpretations of the image that can significantly compromise downstream reasoning or diagnostic accuracy, whereas minor errors involve subtler inaccuracies that do not substantially alter the clinical meaning.
We utilized a curated set of 80 samples (40 from AMBOSS, 40 from MIMIC-CXR) and assessed four models: the base LLaVA-Med, and its SFT baseline, Text-Hallu, and MMedPO. 
More details are provided in Appendix~\ref{appendix:exp_detail}.

Table~\ref{tab:expert_eval} shows that the base LLaVA-Med model exhibited substantial limitations, with 82.5\% of image captioning responses and 82.5\% of report generation responses containing severe errors, and an additional 7.5\% and 15\% containing minor errors, respectively.
These results indicate that the baseline model lacks sufficient capability for accurate medical image interpretation.
Moreover, post-training methods such as SFT and DPO were insufficient in addressing this limitation.
In fact, SFT appeared to amplify errors in image captioning, with severe and minor errors increasing to 85\% and 15\%, respectively.
Text-Hallu and MMedPO reduced severe errors in image captioning (to 50\% and 67.5\%), but simultaneously increased minor errors (from 7.5\% to 30\% and 17.5\%, respectively).

\paragraph{Fine-grained error analysis}
We further categorized the errors into four major types:
(1) Modality misidentification (MM): The model incorrectly identifies the imaging modality, such as mistaking a pathology slide for a clinical photograph.
(2) Spatial or laterality confusion (SLC): The model confuses spatial orientation or left/right anatomical sides, for instance, describing a left lung lesion as being located in the right lung.
(3) Anatomical misidentification (AM): The model misidentifies anatomical structures, such as referring to the lip as intraoral tissue or confusing the arm with the leg.
(4) Lack of anatomical specificity (LAS): The model provides overly broad anatomical references relative to the ground truth, such as describing the ``right lung'' instead of the more precise ``right lower lobe.''
These error types reflect basic visual understanding that should be correctly recognized before any detailed reasoning.
However, as shown in Table~\ref{tab:expert_eval}, models frequently made such errors, and post-training methods even amplified specific categories, for example, AM in image captioning and LAS in report generation.




\begin{table}[t!]
\centering
\small
\setlength{\tabcolsep}{5pt}
\begin{tabular}{>{\raggedright\arraybackslash}p{1.8cm} c cccc}
\toprule
 & \multicolumn{1}{c}{\multirow{2}{*}[-0.6ex]{\shortstack[c]{\textbf{VQA}\\[1pt]\textbf{(Pooled)}}}}
 & \multicolumn{4}{c}{\textbf{Subsets}} \\
\cmidrule(lr){3-6}
\textbf{Model} &  & \textbf{MM} & \textbf{SLC} & \textbf{AM} & \textbf{LAS} \\
\midrule
Base model          & 38.7 & 49.8 & 26.6 & 40.3 & 10.0 \\
SFT                       & 41.0 & 56.2 & 30.2 & 43.2 & 14.2 \\
\midrule
Text-Hallu               & 41.3 & 63.6 & 30.4 & 42.2 & 13.2 \\
\hspace*{0.9em}+ NLL      & 42.0 & 60.1 & 31.2 & 43.0 & 14.9 \\
Text-Noise                & 39.0 & 54.2 & 29.8 & 41.6 & 14.8 \\
\hspace*{0.9em}+ NLL      & 40.6 & 56.0 & 32.3 & 42.7 & 13.7 \\
IRPO                      & 39.0 & 54.1 & 31.8 & 40.8 & 13.9 \\
\midrule
Image-Noise                & 40.0 & 51.8 & 32.2 & 43.4 & 14.9 \\
Image-ROI               & 41.0 & 55.5 & 30.3 & 42.6 & 11.7 \\
\midrule
mDPO                      & 41.9 & 59.5 & 32.1 & 42.7 & 14.7 \\
MMedPO                    & 40.1 & 56.7 & 32.9 & 41.5 & 12.3 \\
\midrule
Ours (DPO)               & 45.5 & \textbf{69.4} & 35.9 & 45.5 & 20.8 \\
Ours (CoPO)              & \textbf{45.6} & 69.0 & \textbf{36.0} & \textbf{45.6} & 20.8 \\
Ours (mDPO)              & 45.4 & 69.1 & 35.9 & 45.5 & \textbf{20.9} \\
\bottomrule
\end{tabular}
\caption{
Performance comparison of our DPO models and baseline models on VQA tasks. 
The left group represents pooled VQA performance (averaged across the full SLAKE, VQA-RAD, and PathVQA datasets). 
The right group shows accuracy on error-specific subsets, consisting only of VQA items relevant to specific visual recognition errors.
MM: Modality misidentification. SLC: Spatial or laterality confusion.
AM: Anatomical misidentification
LAS: Lack of anatomical specificity.}
\label{tab:error_type_eval_extended}
\end{table}

\section{Enhanced DPO Training}
\label{sec:error_specific}

Manual analysis revealed that the majority of responses contained image misinterpretation errors.
Since accurate visual understanding is the foundation for downstream reasoning, such errors often lead to cascading failures in subsequent steps, likely contributing to the overall inconsistency in performance.
Unfortunately, existing DPO methods are not well-equipped to address these underlying limitations of the base models.
Even MMedPO, which was specifically designed to enhance visual grounding by aligning model attention with clinically critical ROIs in medical images, exhibited significant visual errors in our evaluation.
To examine whether this issue is addressable, we explored a straightforward approach by incorporating fine-grained visual error types into the DPO pair construction process. As a proof of concept, we integrated four common categories of model errors (i.e., MM, SLC, AM, and LAS).

\subsection{Model}
We constructed preference pairs that isolate a single error type while preserving the surrounding clinical context across both text and image modalities. 
To ensure a fair comparison, we aggregated samples across all error categories to form a single training set of 10k samples, consistent with the size of the dataset used for our baseline models.
Using this dataset, we developed and evaluated three distinct DPO configurations to investigate the impact of each modality: text-only (DPO), image-only (CoPO), and joint text-image (mDPO). 

\paragraph{Error-type assignment} 
To systematically identify relevant samples, we first defined keyword lists corresponding to each error category (see Appendix~\ref{appendix:keywords}).
These lists were then used to map each image to its potential failure modes by identifying relevant clinical terms within the associated instruction and response.
Since each sample only pertains to specific error types depending on its content, we tagged each image with only the detected categories.
For instance, as illustrated in Figure~\ref{fig:main_fig}c, an image of a chest CT might be tagged with MM: \{CT\}, AM: \{lung\}, and SLC: \{right\}, while LAS is omitted as no corresponding keywords were matched. 
These tags serve as ground-truth anchors, allowing us to formulate contrastive pairs by systematically perturbing specific attributes or selectively retrieving images that align with these anchors, all while keeping the rest of the clinical context intact.

\paragraph{Generation of rejected text responses}
We utilized GPT-4o to generate a corresponding rejected response $y_l$ by perturbing the identified error-type keywords.
Following the SLC example in Figure~\ref{fig:main_fig}c, if the response $y$ describes a finding on the ``right,'' the model was prompted to substitute the target keyword with a plausible but clinically incorrect alternative, resulting in the term ``left'' in the rejected output $y_l$. 
We constrained the generation process to strictly preserve all other clinical details and maintain a consistent length and tone between $y_w$ and  $y_l$.

\paragraph{Retrieval of rejected images}
The original image, serving as the chosen image $m_w$, was paired with a hard-negative $m_l$. 
We selected $m_l$ by retrieving a sample that differed strictly on the targeted attribute; for instance, as shown in Figure~\ref{fig:main_fig}c, we selected an image displaying a ``left'' side pathology instead of ``right'' while keeping other attributes consistent. 
This selection process ensured precise supervision by forcing the model to distinguish subtle spatial differences between otherwise nearly identical clinical contexts in images.

\subsection{Results}
We evaluated our models across three VQA datasets using average accuracy as the primary metric. To conduct a more granular analysis, we applied the same automated classification logic used in our dataset construction to categorize the original questions into the four error types.

As shown in Table~\ref{tab:error_type_eval_extended}, the baseline DPO models yielded inconsistent improvements; while they occasionally surpassed SFT in specific categories, they frequently fell short in others.
In contrast, our proposed models consistently achieved the highest accuracy across all categories, outperforming both SFT and all other DPO variants. This robust performance demonstrates that targeted preference learning on clinically grounded error types provides more reliable hallucination mitigation than general-purpose DPO strategies.

\section{Conclusion}
In this work, we examined the efficacy of existing DPO strategies within the medical domain. Our results demonstrated that while DPO provides moderate gains over the base model, these improvements are often indistinguishable from those of matched SFT baselines, suggesting that the benefits may stem primarily from additional supervised training rather than preference optimization itself. Furthermore, we found that DPO's performance was highly inconsistent across various tasks and datasets, raising concerns regarding its reliability in clinical applications.
Most importantly, our expert-driven analysis of visual understanding revealed that DPO-aligned models still exhibit fundamental errors in modality and anatomical identification.
These results underscore the limitations of current approaches and call for more advanced, domain-specific methods that prioritize visual grounding. 
To this end, we provided a proof-of-concept for a targeted preference construction strategy and released our training and evaluation resources to support the development of more robust medical AI.

\section*{Limitations}

It remains to be verified whether our findings generalize to more recently released models, such as HealthGPT~\cite{lin2025healthgpt} or MedGemma~\cite{sellergren2025medgemma}. 
Variations in base architectures and the scale of medical pre-training data across these newer models may influence how they respond to preference optimization.
Our evaluation was also constrained to 7B-parameter models, primarily due to computational limitations; however, investigating the scaling effects of preference optimization on larger architectures would be a valuable direction.
Furthermore, several specialized models have been developed for radiology report generation~\cite{wu2023towards, chen2024chexagent, lee2024llm}, and evaluating such radiology-specific architectures might have revealed different error patterns.
However, as our primary objective was to assess widely adopted models for general medicine, we selected LLaVA-Med and HuatuoGPT-Vision. 
Also, they represent widely used, open-source benchmarks with transparent training pipelines. 
This transparency was crucial for ensuring that our comparisons across DPO variants remained controlled and reproducible.

While our analysis primarily focused on visual-level errors, investigating failure modes that occur despite accurate visual recognition remains an intriguing avenue for future research. This includes challenges such as extrinsic hallucinations, overly generic descriptions, logical inconsistency in reasoning, and the degree of alignment with global medical standards. We leave the exploration of these nuanced linguistic and clinical dimensions for future work.

Lastly, our study may not comprehensively cover the full range of real-world clinical scenarios. 
As such, various types of errors may arise in practical settings that were not captured or analyzed within the scope of this research.
Therefore, ongoing efforts toward additional validation are necessary to ensure robustness and reliability in diverse medical contexts.

\section*{Acknowledgments}

This research was supported by (1) the National Research Foundation of Korea (NRF-2023R1A2C3004176), (2) the Ministry of Health \& Welfare, Republic of Korea (HR20C002103), (3) ICT Creative Consilience Program through the Institute of Information \& Communications Technology Planning \& Evaluation (IITP) grant funded by the Korea government (MSIT) (IITP-2025-RS-2020-II201819), (4) the National Research Foundation of Korea(NRF) grant funded by the Korea governmant(MSIT and MOE) (RS-2025-16652968), and (5) the Seoul National University Hospital with support from the Ministry of Science and ICT (RS-2023-00262002)


\bibliography{custom}

@article{liu2023visual,
  title={Visual instruction tuning},
  author={Liu, Haotian and Li, Chunyuan and Wu, Qingyang and Lee, Yong Jae},
  journal={Advances in neural information processing systems},
  volume={36},
  pages={34892--34916},
  year={2023}
}

@article{li2023llava,
  title={Llava-med: Training a large language-and-vision assistant for biomedicine in one day},
  author={Li, Chunyuan and Wong, Cliff and Zhang, Sheng and Usuyama, Naoto and Liu, Haotian and Yang, Jianwei and Naumann, Tristan and Poon, Hoifung and Gao, Jianfeng},
  journal={Advances in Neural Information Processing Systems},
  volume={36},
  pages={28541--28564},
  year={2023}
}

@inproceedings{chen2024huatuogpt,
  title={Towards injecting medical visual knowledge into multimodal llms at scale},
  author={Chen, Junying and Gui, Chi and Ouyang, Ruyi and Gao, Anningzhe and Chen, Shunian and Chen, Guiming Hardy and Wang, Xidong and Cai, Zhenyang and Ji, Ke and Wan, Xiang and others},
  booktitle={Proceedings of the 2024 conference on empirical methods in natural language processing},
  pages={7346--7370},
  year={2024}
}

@article{rafailov2023direct,
  title={Direct preference optimization: Your language model is secretly a reward model},
  author={Rafailov, Rafael and Sharma, Archit and Mitchell, Eric and Manning, Christopher D and Ermon, Stefano and Finn, Chelsea},
  journal={Advances in Neural Information Processing Systems},
  volume={36},
  pages={53728--53741},
  year={2023}
}

@article{bai2024hallucination,
  title={Hallucination of multimodal large language models: A survey},
  author={Bai, Zechen and Wang, Pichao and Xiao, Tianjun and He, Tong and Han, Zongbo and Zhang, Zheng and Shou, Mike Zheng},
  journal={arXiv preprint arXiv:2404.18930},
  year={2024}
}

@article{zhao2023beyond,
  title={Beyond hallucinations: Enhancing lvlms through hallucination-aware direct preference optimization},
  author={Zhao, Zhiyuan and Wang, Bin and Ouyang, Linke and Dong, Xiaoyi and Wang, Jiaqi and He, Conghui},
  journal={arXiv preprint arXiv:2311.16839},
  year={2023}
}

@inproceedings{xiao2025detecting,
  title={Detecting and mitigating hallucination in large vision language models via fine-grained ai feedback},
  author={Xiao, Wenyi and Huang, Ziwei and Gan, Leilei and He, Wanggui and Li, Haoyuan and Yu, Zhelun and Shu, Fangxun and Jiang, Hao and Zhu, Linchao},
  booktitle={Proceedings of the AAAI Conference on Artificial Intelligence},
  volume={39},
  number={24},
  pages={25543--25551},
  year={2025}
}

@article{li2023silkie,
  title={Silkie: Preference distillation for large visual language models},
  author={Li, Lei and Xie, Zhihui and Li, Mukai and Chen, Shunian and Wang, Peiyi and Chen, Liang and Yang, Yazheng and Wang, Benyou and Kong, Lingpeng},
  journal={arXiv preprint arXiv:2312.10665},
  year={2023}
}

@inproceedings{zhu2024mmedpo,
  title={MMedPO: Aligning Medical Vision-Language Models with Clinical-Aware Multimodal Preference Optimization},
  author={Zhu, Kangyu and Xia, Peng and Li, Yun and Zhu, Hongtu and Wang, Sheng and Yao, Huaxiu},
  booktitle={Forty-second International Conference on Machine Learning},
  year={2025}
}

@article{zhou2024aligning,
  title={Aligning modalities in vision large language models via preference fine-tuning},
  author={Zhou, Yiyang and Cui, Chenhang and Rafailov, Rafael and Finn, Chelsea and Yao, Huaxiu},
  journal={arXiv preprint arXiv:2402.11411},
  year={2024}
}

@inproceedings{wang2025enhancing,
  title={Enhancing visual-language modality alignment in large vision language models via self-improvement},
  author={Wang, Xiyao and Chen, Jiuhai and Wang, Zhaoyang and Zhou, Yuhang and Zhou, Yiyang and Yao, Huaxiu and Zhou, Tianyi and Goldstein, Tom and Bhatia, Parminder and Kass-Hout, Taha and others},
  booktitle={Findings of the Association for Computational Linguistics: NAACL 2025},
  pages={268--282},
  year={2025}
}

@InProceedings{ouali2024clipdpo,
author="Ouali, Yassine
and Bulat, Adrian
and Martinez, Brais
and Tzimiropoulos, Georgios",
editor="Leonardis, Ale{\v{s}}
and Ricci, Elisa
and Roth, Stefan
and Russakovsky, Olga
and Sattler, Torsten
and Varol, G{\"u}l",
title="CLIP-DPO: Vision-Language Models as a Source of Preference for Fixing Hallucinations in LVLMs",
booktitle="Computer Vision -- ECCV 2024",
year="2025",
publisher="Springer Nature Switzerland",
address="Cham",
pages="395--413",
isbn="978-3-031-73116-7"
}

@inproceedings{wang2024mdpo,
  title={mDPO: Conditional Preference Optimization for Multimodal Large Language Models},
  author={Wang, Fei and Zhou, Wenxuan and Huang, James Y and Xu, Nan and Zhang, Sheng and Poon, Hoifung and Chen, Muhao},
  booktitle={Proceedings of the 2024 Conference on Empirical Methods in Natural Language Processing},
  pages={8078--8088},
  year={2024}
}

@article{deng2024enhancing,
  title={Enhancing large vision language models with self-training on image comprehension},
  author={Deng, Yihe and Lu, Pan and Yin, Fan and Hu, Ziniu and Shen, Sheng and Gu, Quanquan and Zou, James Y and Chang, Kai-Wei and Wang, Wei},
  journal={Advances in Neural Information Processing Systems},
  volume={37},
  pages={131369--131397},
  year={2024}
}

@article{chen2024medhallmark,
  title={Detecting and evaluating medical hallucinations in large vision language models},
  author={Chen, Jiawei and Yang, Dingkang and Wu, Tong and Jiang, Yue and Hou, Xiaolu and Li, Mingcheng and Wang, Shunli and Xiao, Dongling and Li, Ke and Zhang, Lihua},
  journal={arXiv preprint arXiv:2406.10185},
  year={2024}
}

@article{chang2025medheval,
  title={MedHEval: Benchmarking Hallucinations and Mitigation Strategies in Medical Large Vision-Language Models},
  author={Chang, Aofei and Huang, Le and Bhatia, Parminder and Kass-Hout, Taha and Ma, Fenglong and Xiao, Cao},
  journal={arXiv preprint arXiv:2503.02157},
  year={2025}
}

@article{pang2024iterative,
  title={Iterative reasoning preference optimization},
  author={Pang, Richard Yuanzhe and Yuan, Weizhe and He, He and Cho, Kyunghyun and Sukhbaatar, Sainbayar and Weston, Jason},
  journal={Advances in Neural Information Processing Systems},
  volume={37},
  pages={116617--116637},
  year={2024}
}

@article{lau2018dataset,
  title={A dataset of clinically generated visual questions and answers about radiology images},
  author={Lau, Jason J and Gayen, Soumya and Ben Abacha, Asma and Demner-Fushman, Dina},
  journal={Scientific data},
  volume={5},
  number={1},
  pages={1--10},
  year={2018},
  publisher={Nature Publishing Group}
}

@inproceedings{liu2021slake,
  title={Slake: A semantically-labeled knowledge-enhanced dataset for medical visual question answering},
  author={Liu, Bo and Zhan, Li-Ming and Xu, Li and Ma, Lin and Yang, Yan and Wu, Xiao-Ming},
  booktitle={2021 IEEE 18th international symposium on biomedical imaging (ISBI)},
  pages={1650--1654},
  year={2021},
  organization={IEEE}
}

@article{johnson2019mimic,
  title={MIMIC-CXR, a de-identified publicly available database of chest radiographs with free-text reports},
  author={Johnson, Alistair EW and Pollard, Tom J and Berkowitz, Seth J and Greenbaum, Nathaniel R and Lungren, Matthew P and Deng, Chih-ying and Mark, Roger G and Horng, Steven},
  journal={Scientific data},
  volume={6},
  number={1},
  pages={317},
  year={2019},
  publisher={Nature Publishing Group UK London}
}

@inproceedings{wu2023medklip,
  title={Medklip: Medical knowledge enhanced language-image pre-training for x-ray diagnosis},
  author={Wu, Chaoyi and Zhang, Xiaoman and Zhang, Ya and Wang, Yanfeng and Xie, Weidi},
  booktitle={Proceedings of the IEEE/CVF International Conference on Computer Vision},
  pages={21372--21383},
  year={2023}
}

@inproceedings{xie2024medtrinity,
  title={MedTrinity-25M: A Large-scale Multimodal Dataset with Multigranular Annotations for Medicine},
  author={Xie, Yunfei and Zhou, Ce and Gao, Lang and Wu, Juncheng and Li, Xianhang and Zhou, Hong-Yu and Liu, Sheng and Xing, Lei and Zou, James and Xie, Cihang and others},
  booktitle={The Thirteenth International Conference on Learning Representations},
  year={2025}
}

@article{wu2023towards,
  title={Towards generalist foundation model for radiology by leveraging web-scale 2d\&3d medical data},
  author={Wu, Chaoyi and Zhang, Xiaoman and Zhang, Ya and Hui, Hui and Wang, Yanfeng and Xie, Weidi},
  journal={Nature Communications},
  volume={16},
  number={1},
  pages={7866},
  year={2025},
  publisher={Nature Publishing Group UK London}
}

@article{openai2023gpt4v,
 author = {OpenAI},
 title = {GPT-4V(ision) system card},
 year = {2023}
}

@article{alayrac2022flamingo,
  title={Flamingo: a visual language model for few-shot learning},
  author={Alayrac, Jean-Baptiste and Donahue, Jeff and Luc, Pauline and Miech, Antoine and Barr, Iain and Hasson, Yana and Lenc, Karel and Mensch, Arthur and Millican, Katherine and Reynolds, Malcolm and others},
  journal={Advances in neural information processing systems},
  volume={35},
  pages={23716--23736},
  year={2022}
}

@inproceedings{li2023blip,
  title={Blip-2: Bootstrapping language-image pre-training with frozen image encoders and large language models},
  author={Li, Junnan and Li, Dongxu and Savarese, Silvio and Hoi, Steven},
  booktitle={International conference on machine learning},
  pages={19730--19742},
  year={2023},
  organization={PMLR}
}

@inproceedings{zhu2023minigpt,
  title={MINIGPT-4: ENHANCING VISION-LANGUAGE UNDERSTANDING WITH ADVANCED LARGE LANGUAGE MODELS},
  author={Zhu, Deyao and Chen, Jun and Shen, Xiaoqian and Li, Xiang and Elhoseiny, Mohamed},
  booktitle={12th International Conference on Learning Representations, ICLR 2024},
  year={2024}
}

@article{kline2022multimodal,
  title={Multimodal machine learning in precision health: A scoping review},
  author={Kline, Adrienne and Wang, Hanyin and Li, Yikuan and Dennis, Saya and Hutch, Meghan and Xu, Zhenxing and Wang, Fei and Cheng, Feixiong and Luo, Yuan},
  journal={npj Digital Medicine},
  volume={5},
  number={1},
  pages={171},
  year={2022},
  publisher={Nature Publishing Group UK London}
}

@article{jin2024hidden,
  title={Hidden flaws behind expert-level accuracy of multimodal GPT-4 vision in medicine},
  author={Jin, Qiao and Chen, Fangyuan and Zhou, Yiliang and Xu, Ziyang and Cheung, Justin M and Chen, Robert and Summers, Ronald M and Rousseau, Justin F and Ni, Peiyun and Landsman, Marc J and others},
  journal={npj Digital Medicine},
  volume={7},
  number={1},
  pages={190},
  year={2024},
  publisher={Nature Publishing Group UK London}
}

@article{liu2024survey,
  title={A survey on hallucination in large vision-language models},
  author={Liu, Hanchao and Xue, Wenyuan and Chen, Yifei and Chen, Dapeng and Zhao, Xiutian and Wang, Ke and Hou, Liping and Li, Rongjun and Peng, Wei},
  journal={arXiv preprint arXiv:2402.00253},
  year={2024}
}

@inproceedings{maynez2020faithfulness,
  title={On Faithfulness and Factuality in Abstractive Summarization},
  author={Maynez, Joshua and Narayan, Shashi and Bohnet, Bernd and McDonald, Ryan},
  booktitle={Proceedings of the 58th Annual Meeting of the Association for Computational Linguistics},
  pages={1906--1919},
  year={2020}
}

@article{kim2025medical,
  title={Medical Hallucination in Foundation Models and Their Impact on Healthcare},
  author={Kim, Yubin and Jeong, Hyewon and Chen, Shen and Li, Shuyue Stella and Lu, Mingyu and Alhamoud, Kumail and Mun, Jimin and Grau, Cristina and Jung, Minseok and Gameiro, Rodrigo R and others},
  journal={medRxiv},
  pages={2025--02},
  year={2025},
  publisher={Cold Spring Harbor Laboratory Press}
}

@article{he2020pathvqa,
  title={Pathvqa: 30000+ questions for medical visual question answering},
  author={He, Xuehai and Zhang, Yichen and Mou, Luntian and Xing, Eric and Xie, Pengtao},
  journal={arXiv preprint arXiv:2003.10286},
  year={2020}
}

@inproceedings{min2023factscore,
  title={Factscore: Fine-grained atomic evaluation of factual precision in long form text generation},
  author={Min, Sewon and Krishna, Kalpesh and Lyu, Xinxi and Lewis, Mike and Yih, Wen-tau and Koh, Pang and Iyyer, Mohit and Zettlemoyer, Luke and Hajishirzi, Hannaneh},
  booktitle={Proceedings of the 2023 Conference on Empirical Methods in Natural Language Processing},
  pages={12076--12100},
  year={2023}
}

@inproceedings{zhang2024language,
  title={How Language Model Hallucinations Can Snowball},
  author={Zhang, Muru and Press, Ofir and Merrill, William and Liu, Alisa and Smith, Noah A},
  booktitle={International Conference on Machine Learning},
  pages={59670--59684},
  year={2024},
  organization={PMLR}
}

@article{zhang2024generalist,
  title={A generalist vision--language foundation model for diverse biomedical tasks},
  author={Zhang, Kai and Zhou, Rong and Adhikarla, Eashan and Yan, Zhiling and Liu, Yixin and Yu, Jun and Liu, Zhengliang and Chen, Xun and Davison, Brian D and Ren, Hui and others},
  journal={Nature Medicine},
  pages={1--13},
  year={2024},
  publisher={Nature Publishing Group US New York}
}

@inproceedings{chen2024chexagent,
  title={Chexagent: Towards a foundation model for chest x-ray interpretation},
  author={Chen, Zhihong and Varma, Maya and Delbrouck, Jean-Benoit and Paschali, Magdalini and Blankemeier, Louis and Van Veen, Dave and Valanarasu, Jeya Maria Jose and Youssef, Alaa and Cohen, Joseph Paul and Reis, Eduardo Pontes and others},
  booktitle={AAAI 2024 Spring Symposium on Clinical Foundation Models},
  year={2024}
}

@inproceedings{lee2024llm,
  title={LLM-CXR: Instruction-Finetuned LLM for CXR Image Understanding and Generation},
  author={Lee, Suhyeon and Kim, Won Jun and Chang, Jinho and Ye, Jong Chul},
  booktitle={The Twelfth International Conference on Learning Representations},
  year={2024}
}

@inproceedings{wang2024surgical,
  title={Surgical-LVLM: Learning to Adapt Large Vision-Language Model for Grounded Visual Question Answering in Robotic Surgery},
  author={Wang, Guankun and Bai, Long and Nah, Wan Jun and Wang, Jie and Zhang, Zhaoxi and Chen, Zhen and Wu, Jinlin and Islam, Mobarakol and Liu, Hongbin and Ren, Hongliang},
  booktitle={ICLR 2025 Workshop on Foundation Models in the Wild},
  year={2025}
}

@article{zhang2025multimodal,
  title={A multimodal biomedical foundation model trained from fifteen million image--text pairs},
  author={Zhang, Sheng and Xu, Yanbo and Usuyama, Naoto and Xu, Hanwen and Bagga, Jaspreet and Tinn, Robert and Preston, Sam and Rao, Rajesh and Wei, Mu and Valluri, Naveen and others},
  journal={NEJM AI},
  volume={2},
  number={1},
  pages={AIoa2400640},
  year={2025},
  publisher={Massachusetts Medical Society}
}

@inproceedings{lin2023pmc,
  title={Pmc-clip: Contrastive language-image pre-training using biomedical documents},
  author={Lin, Weixiong and Zhao, Ziheng and Zhang, Xiaoman and Wu, Chaoyi and Zhang, Ya and Wang, Yanfeng and Xie, Weidi},
  booktitle={International Conference on Medical Image Computing and Computer-Assisted Intervention},
  pages={525--536},
  year={2023},
  organization={Springer}
}

@article{zhou2024pre,
  title={Pre-trained multimodal large language model enhances dermatological diagnosis using SkinGPT-4},
  author={Zhou, Juexiao and He, Xiaonan and Sun, Liyuan and Xu, Jiannan and Chen, Xiuying and Chu, Yuetan and Zhou, Longxi and Liao, Xingyu and Zhang, Bin and Afvari, Shawn and others},
  journal={Nature Communications},
  volume={15},
  number={1},
  pages={5649},
  year={2024},
  publisher={Nature Publishing Group UK London}
}

@inproceedings{seyfioglu2024quilt,
  title={Quilt-llava: Visual instruction tuning by extracting localized narratives from open-source histopathology videos},
  author={Seyfioglu, Mehmet Saygin and Ikezogwo, Wisdom O and Ghezloo, Fatemeh and Krishna, Ranjay and Shapiro, Linda},
  booktitle={Proceedings of the IEEE/CVF Conference on Computer Vision and Pattern Recognition},
  pages={13183--13192},
  year={2024}
}

@inproceedings{royer2024multimedeval,
  title={MultiMedEval: A Benchmark and a Toolkit for Evaluating Medical Vision-Language Models},
  author={Royer, Corentin and Menze, Bjoern and Sekuboyina, Anjany},
  booktitle={Medical Imaging with Deep Learning},
  pages={1310--1327},
  year={2024},
  organization={PMLR}
}

@article{liu2024spectrum,
  title={A Spectrum Evaluation Benchmark for Medical Multi-Modal Large Language Models},
  author={Liu, Jie and Wang, Wenxuan and Su, Yihang and Huan, Jingyuan and Chen, Wenting and Zhang, Yudi and Li, Cheng-Yi and Chang, Kao-Jung and Xin, Xiaohan and Shen, Linlin and others},
  journal={arXiv preprint arXiv:2402.11217},
  year={2024}
}

@article{xia2024cares,
  title={Cares: A comprehensive benchmark of trustworthiness in medical vision language models},
  author={Xia, Peng and Chen, Ze and Tian, Juanxi and Gong, Yangrui and Hou, Ruibo and Xu, Yue and Wu, Zhenbang and Fan, Zhiyuan and Zhou, Yiyang and Zhu, Kangyu and others},
  journal={Advances in Neural Information Processing Systems},
  volume={37},
  pages={140334--140365},
  year={2024}
}

@inproceedings{saeidi2025insights,
  title={Insights into alignment: Evaluating dpo and its variants across multiple tasks},
  author={Saeidi, Amir and Verma, Shivanshu and Uddin, Md Nayem and Baral, Chitta},
  booktitle={Proceedings of the 63rd Annual Meeting of the Association for Computational Linguistics (Volume 4: Student Research Workshop)},
  pages={409--421},
  year={2025}
}

@inproceedings{xu2024contrastive,
  title={Contrastive preference optimization: pushing the boundaries of LLM performance in machine translation},
  author={Xu, Haoran and Sharaf, Amr and Chen, Yunmo and Tan, Weiting and Shen, Lingfeng and Van Durme, Benjamin and Murray, Kenton and Kim, Young Jin},
  booktitle={Proceedings of the 41st International Conference on Machine Learning},
  pages={55204--55224},
  year={2024}
}

@article{ethayarajh2024kto,
  title={Kto: Model alignment as prospect theoretic optimization},
  author={Ethayarajh, Kawin and Xu, Winnie and Muennighoff, Niklas and Jurafsky, Dan and Kiela, Douwe},
  journal={arXiv preprint arXiv:2402.01306},
  year={2024}
}

@article{zheng2023judging,
  title={Judging llm-as-a-judge with mt-bench and chatbot arena},
  author={Zheng, Lianmin and Chiang, Wei-Lin and Sheng, Ying and Zhuang, Siyuan and Wu, Zhanghao and Zhuang, Yonghao and Lin, Zi and Li, Zhuohan and Li, Dacheng and Xing, Eric and others},
  journal={Advances in neural information processing systems},
  volume={36},
  pages={46595--46623},
  year={2023}
}

@article{gu2024survey,
  title={A survey on llm-as-a-judge},
  author={Gu, Jiawei and Jiang, Xuhui and Shi, Zhichao and Tan, Hexiang and Zhai, Xuehao and Xu, Chengjin and Li, Wei and Shen, Yinghan and Ma, Shengjie and Liu, Honghao and others},
  journal={The Innovation},
  year={2024},
  publisher={Elsevier}
}

@inproceedings{lin2025healthgpt,
  title={HealthGPT: A Medical Large Vision-Language Model for Unifying Comprehension and Generation via Heterogeneous Knowledge Adaptation},
  author={Lin, Tianwei and Zhang, Wenqiao and LI, SIJING and Yuan, Yuqian and Yu, Binhe and Li, Haoyuan and He, Wanggui and Jiang, Hao and Li, Mengze and Tang, Siliang and others},
  booktitle={Forty-second International Conference on Machine Learning},
  year={2025}
}

@article{sellergren2025medgemma,
  title={Medgemma technical report},
  author={Sellergren, Andrew and Kazemzadeh, Sahar and Jaroensri, Tiam and Kiraly, Atilla and Traverse, Madeleine and Kohlberger, Timo and Xu, Shawn and Jamil, Fayaz and Hughes, C{\'\i}an and Lau, Charles and others},
  journal={arXiv preprint arXiv:2507.05201},
  year={2025}
}

\clearpage

\appendix

\renewcommand{\thetable}{\Alph{table}}
\renewcommand{\thefigure}{\Alph{figure}}
\setcounter{table}{0}
\setcounter{figure}{0}

\begin{table*}[t]
  \footnotesize
  \centering
  \renewcommand{\arraystretch}{1.6}
  \begin{tabular}{>{\raggedright\arraybackslash}p{4cm}
                  >{\raggedright\arraybackslash}p{11cm}}
    \toprule
    \footnotesize\textbf {Method} & \footnotesize\textbf {Objective} \\
    \midrule
        \raggedright IRPO \cite{pang2024iterative} &
    \parbox[t]{12cm}{
      \raggedright
      $\begin{array}{@{}l@{}}
        \mathcal{L}_{\text{IRPO}}(y_w, y_l \mid m, q) = \mathcal{L}_{\text{DPO}}(y_w, y_l \mid m, q) + \alpha \cdot \mathcal{L}_{\text{NLL}}(y_w \mid m, q) \\
        = - \log \sigma \Big( \beta \log \tfrac{\pi_\theta(y_w \mid m, q)}{\pi_{\mathrm{ref}}(y_w \mid m, q)} - 
        \beta \log \tfrac{\pi_\theta(y_l \mid m, q)}{\pi_{\mathrm{ref}}(y_l \mid m, q)} \Big) - \alpha \cdot \tfrac{\log \pi_\theta(y_w \mid m, q)}{|y_w|}
      \end{array}$
    } \\
    \midrule
    \raggedright mDPO \cite{wang2024mdpo} &
    \parbox[t]{12cm}{
      \raggedright
      $\begin{array}{@{}l@{}}
        \mathcal{L}_{\text{mDPO}} = \mathcal{L}_{\text{DPO}_m} + \mathcal{L}_{\text{CoPO}} + \mathcal{L}_{\text{AncPO}} \\
        = - \log \sigma \Big( \beta \log \tfrac{\pi_\theta(y_w \mid m, q)}{\pi_{\mathrm{ref}}(y_w \mid m, q)} - 
        \beta \log \tfrac{\pi_\theta(y_l \mid m, q)}{\pi_{\mathrm{ref}}(y_l \mid m, q)} \Big) \\
        \quad - \log \sigma \Big( \beta \log \tfrac{\pi_\theta(y_w \mid m_w, q)}{\pi_{\mathrm{ref}}(y_w \mid m_w, q)} - 
        \beta \log \tfrac{\pi_\theta(y_w \mid m_l, q)}{\pi_{\mathrm{ref}}(y_w \mid m_l, q)} \Big) \\
        \quad - \log \sigma \Big( \beta \log \tfrac{\pi_\theta(y_w \mid m_w, q)}{\pi_{\mathrm{ref}}(y_w \mid m_w, q)} - \delta \Big)
      \end{array}$
    } \\
    \midrule
    \raggedright MMedPO \cite{zhu2024mmedpo} &
    \parbox[t]{12cm}{
      \raggedright
      $\begin{array}{@{}l@{}}
        \mathcal{L}_{\text{MMedPO}} = s' \cdot \Big[
        - \log \sigma \Big(
        \alpha \log \tfrac{\pi_\theta(y_w \mid m_w, q)}{\pi_{\mathrm{ref}}(y_w \mid m_w, q)} - \alpha \log \tfrac{\pi_\theta(y_l \mid m_l, q)}{\pi_{\mathrm{ref}}(y_l \mid m_l, q)}
        \Big) \Big]
      \end{array}$
    } \\
    \bottomrule
  \end{tabular}
    \caption{Mathematical formulations of IRPO, mDPO, and MMedPO.}
  \label{tab:dpo-comparison}
\end{table*}

\section{DPO Formulations and Examples}
\label{sec:appendix}

\begin{figure*}[t!]
  \centering
  \begin{subfigure}{0.45\textwidth}
    \includegraphics[width=\linewidth]{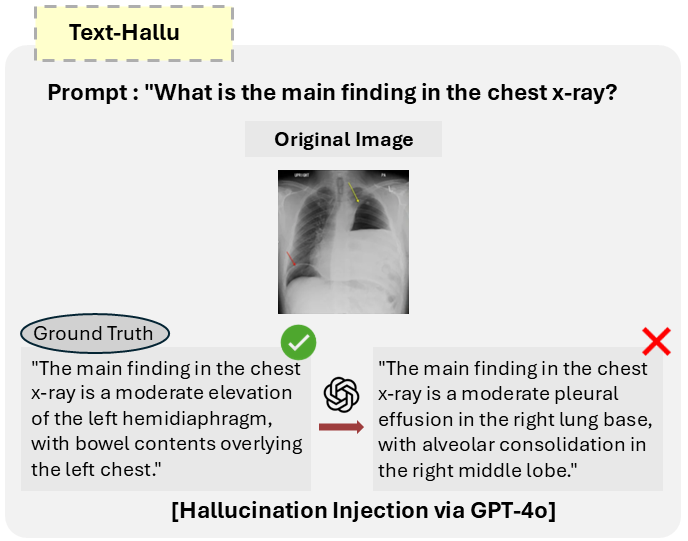}
    \label{fig:TxtPert-LLM}
  \end{subfigure}\hfill
  \begin{subfigure}{0.45\textwidth}
    \includegraphics[width=\linewidth]{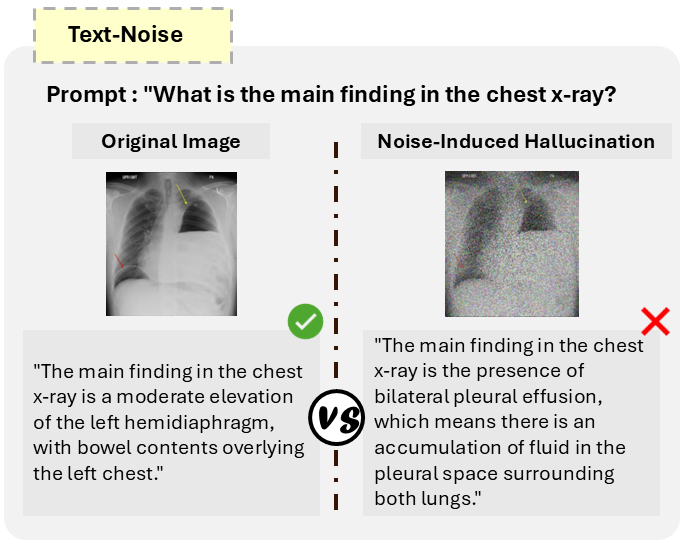}
    \label{fig:TxtPert-GN}
  \end{subfigure}

  \begin{subfigure}{0.45\textwidth}
    \includegraphics[width=\linewidth]{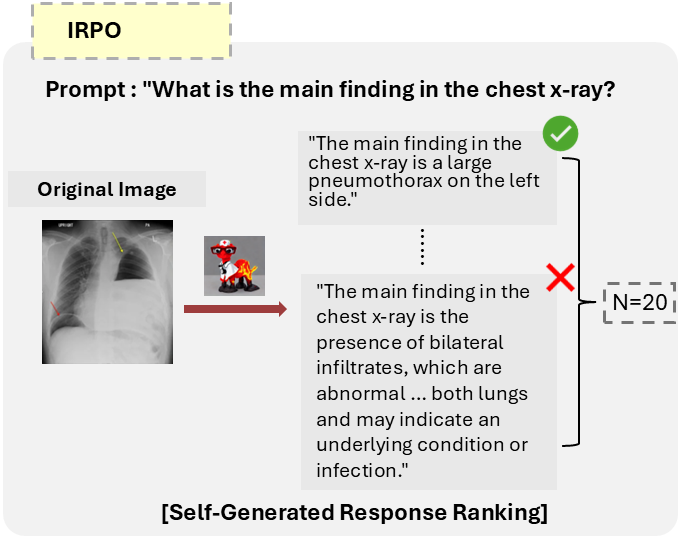}
    \label{fig:IRPO}
  \end{subfigure}\hfill
  \begin{subfigure}{0.45\textwidth}
    \includegraphics[width=\linewidth]{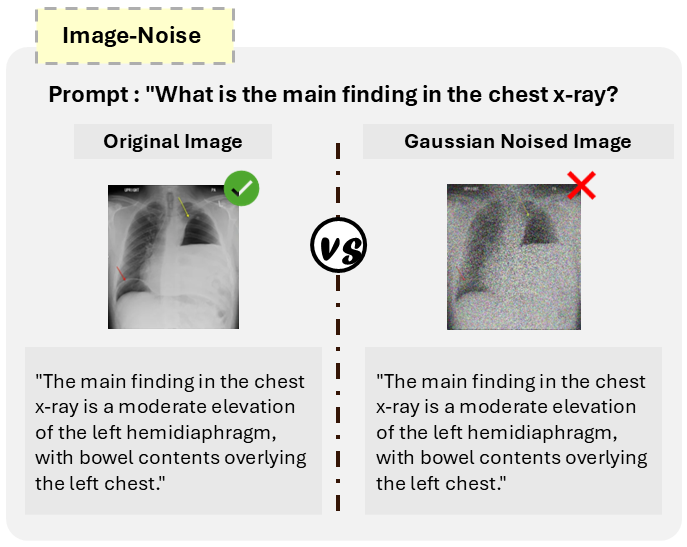}
    \label{fig:ImgPert-GN}
  \end{subfigure}
  \begin{subfigure}{0.45\textwidth}
    \includegraphics[width=\linewidth]{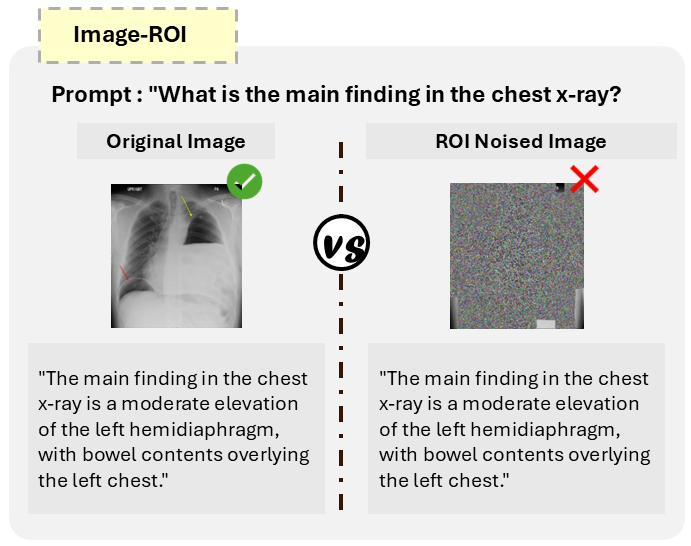}
    \label{fig:ImgPert-ROI}
  \end{subfigure}\hfill
  \hspace{0.45\textwidth}

  \caption{Illustrative examples of preference pair curation in text-only and image-only DPO variants.}
  \label{fig:prefpairs-1}
\end{figure*}

\begin{figure*}[t!]
  \centering
  \begin{subfigure}[t]{0.48\textwidth}
    \centering
    \includegraphics[width=\linewidth]{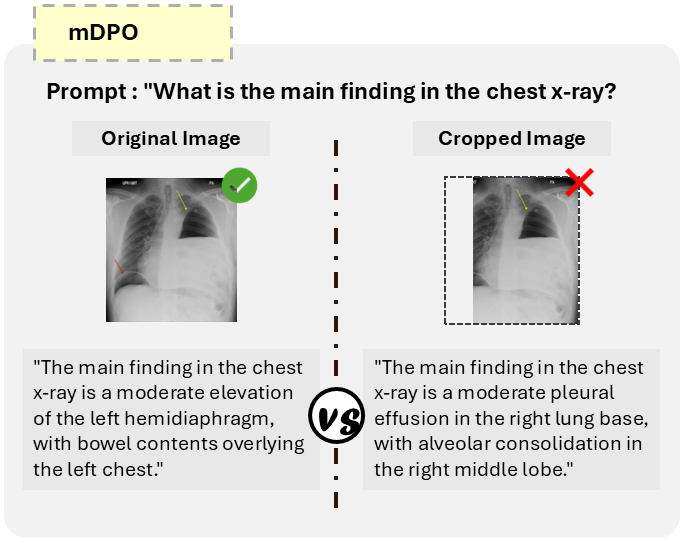}
    \caption{mDPO}
    \label{fig:mDPO}
  \end{subfigure}
  \hfill 
  \begin{subfigure}[t]{0.48\textwidth}
    \centering
    \includegraphics[width=\linewidth]{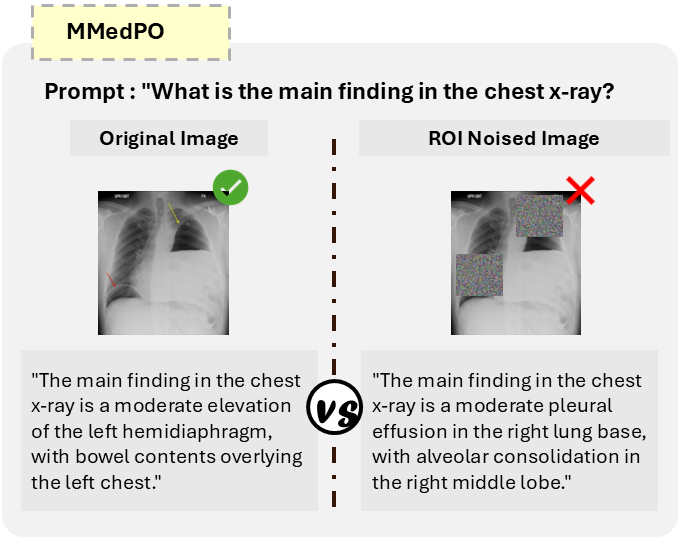}
    \caption{MMedPO}
    \label{fig:MMedPO}
  \end{subfigure}
  
  \caption{Illustrative examples of preference pair curation in joint image-text DPO variants, mDPO and MMedPO.}
  \label{fig:prefpairs-2}
\end{figure*}

Table~\ref{tab:dpo-comparison} presents the mathematical formulations of IRPO, mDPO, and MMedPO, illustrating how the preference loss is defined for each configuration.
Illustrative examples of preference pair curation are presented in Figures~\ref{fig:prefpairs-1} and~\ref{fig:prefpairs-2}.

\section{Completeness and Contradiction}
\label{appendix:nli}

Each model-generated output is evaluated against a set of atomic statements using GPT-4o based natural language inference (NLI) and classified into one of four classes: entailment, if the model's output supports or conveys the same factual content as the reference; partial entailment, if the output is only partially aligned with the statement, capturing some but not all aspects of the intended meaning; contradiction, if the output directly conflicts with the statement; and neutral, if the response neither confirms nor refutes the information, or fails to address it altogether.

Scores of 1, 0.5, 0, and -1 are assigned to entailment, partial entailment, neutral, and contradiction, respectively. Completeness is the average score across the entailment, partial, and neutral categories, while contradiction is the absolute average of the scores for the contradiction class, both normalized by the total number of reference atomic statements.

\section{MIMIC-CXR data curation}
\label{appendix:mimic_curation}

To enable precise and image-grounded evaluation, we utilized the MIMIC-CXR test set after applying the following filters: (1) only studies with a single frontal chest X-ray image were retained; (2) only the Findings section of each report was used, and reports with extremely short Findings sections were excluded due to insufficient clinical content; and (3) we used GPT-4o to generate modified versions of the reports by removing phrases that required external context—such as prior exams, patient history, institutional conventions, or physician-specific commentary.
This selection enables decomposition of reports into atomic, image-verifiable facts for accurate comparison with model outputs.

\section{Expert Evaluation Details}
\label{appendix:exp_detail}

The models were presented with the following prompt:
``Describe the key visual features of the medical image (e.g., shape, size, location, density, contrast). Then, provide the clinical findings.''
Evaluators measured the accuracy of image understanding by assigning one of three severity levels for image misunderstanding:
(1) None: no misinterpretation of critical visual elements, (2) Minor: small inaccuracies that do not substantially affect diagnostic reasoning, and (3) Severe: clear misinterpretation of essential features necessary for accurate clinical inference.

Two annotators with relevant medical backgrounds participated in the expert evaluation. The senior annotator is a licensed physician
specializing in Physical Medicine and Rehabilitation, with years of inpatient experience managing complex comorbidities
and interpreting diverse clinical data, including imaging. The second annotator is a medical student
with prior experience in annotation and manual evaluation across multiple AI projects.
Although our benchmarks span multiple medical domains, the evaluation did not require highly specialized expertise 
from pathologists or radiologists, as the task primarily involved comparing model outputs against available ground truths 
(e.g., radiology reports for MIMIC-CXR and image captions for AMBOSS).

A calibration session was conducted prior to annotation to align evaluation standards. 
To quantify annotation consistency, we computed inter-rater reliability (Cohen's $\kappa$) over 30 model-generated
responses. 
Overall agreement was 0.9 for MIMIC-CXR and was 0.878 for AMBOSS, indicating strong reliability and consensus.

\section{Details of Enhanced DPO Experiments}
\label{appendix:keywords}

\paragraph{Keyword lists}
We curated comprehensive keyword lists for each error category to facilitate automated preference pair construction. 
These keywords serve as the basis for identifying critical clinical entities within the ground-truth instructions and responses. 
The specific keywords associated with each error type are detailed below, illustrating the scope of our targeted clinical entity extraction.

\begin{tcolorbox}[enhanced, breakable, width=\linewidth,
  colback=black!2, colframe=black!35, arc=2pt,
  boxrule=0.3pt, left=6pt, right=6pt, top=6pt, bottom=6pt,
  fontupper=\small\raggedright,
  title=Modality Misidentification (MM)]
\textbf{Representative keywords:}
\begin{itemize}[leftmargin=1.2em, itemsep=2pt]
  \item CT, computed tomography, MRI, MR, T1, T2, FLAIR, DWI, SWI
  \item X-ray, radiograph, CXR, ultrasound, US, sonography, echocardiogram, echo
  \item PET, SPECT, angiography, fluoroscopy, mammography
  \item fundus, ophthalmoscopy, dermatoscopy, endoscopy, colonoscopy, EGD, gastroscopy
  \item microscopy, H\&E, hematoxylin, eosin, electron microscopy, OCT
\end{itemize}
\end{tcolorbox}

\begin{tcolorbox}[enhanced, breakable, width=\linewidth,
  colback=black!2, colframe=black!35, arc=2pt,
  boxrule=0.3pt, left=6pt, right=6pt, top=6pt, bottom=6pt,
  fontupper=\small\raggedright,
  title=Anatomical Misidentification (AM)]
\textbf{Representative keywords (examples):}
\begin{itemize}[leftmargin=1.2em, itemsep=2pt]
  \item \emph{Thorax}: lung, lobe, segment, pleura, mediastinum, cardiomediastinum, diaphragm, rib, clavicle
  \item \emph{Abdomen}: liver, spleen, kidney, adrenal, pancreas, stomach, bowel, colon, rectum
  \item \emph{Head/Neck}: brain, cerebellum, ventricle, skull, orbit, sinus, maxillary, ethmoid, sphenoid, frontal, nasal, septum, tonsil, pharynx, larynx
  \item \emph{Extremities/Skin}: arm, leg, hand, foot, femur, humerus, radius, ulna, tibia, fibula, hip, knee, ankle, wrist, skin, dermis, epidermis
\end{itemize}
\end{tcolorbox}

\begin{tcolorbox}[enhanced, breakable, width=\linewidth,
  colback=black!2, colframe=black!35, arc=2pt,
  boxrule=0.3pt, left=6pt, right=6pt, top=6pt, bottom=6pt,
  fontupper=\small\raggedright,
  title=Spatial or Laterality Confusion (SLC)]
\textbf{Representative keywords:}
\begin{itemize}[leftmargin=1.2em, itemsep=2pt]
  \item Laterality: left, right, left-sided, right-sided
  \item Zones: upper, lower, superior, inferior, anterior, posterior, medial, lateral, apical, basal
  \item Lung subregions: RUL, RML, RLL, LUL, LLL, upper lobe, middle lobe, lower lobe
\end{itemize}
\end{tcolorbox}

\begin{tcolorbox}[enhanced, breakable, width=\linewidth,
  colback=black!2, colframe=black!35, arc=2pt,
  boxrule=0.3pt, left=6pt, right=6pt, top=6pt, bottom=6pt,
  fontupper=\small\raggedright,
  title=Lack of Anatomical Specificity (LAS)]
\textbf{Representative keywords:}
\begin{itemize}[leftmargin=1.2em, itemsep=2pt]
  \item Fine-grained: segment numbers (S1, S2, ...), right lower lobe, left upper lobe, quadrant (RUQ, LUQ, RLQ, LLQ), pole
  \item Broad parents: lung, liver, kidney, sinus, paranasal sinus, brain, hemithorax
\end{itemize}
\end{tcolorbox}

\paragraph{VQA Subsets}
To evaluate the model's robustness against specific types of hallucinations, we constructed specialized evaluation subsets from established VQA benchmarks. By applying the keyword-based classification logic described above, we partitioned original VQA questions into four distinct categories: MM, AM, SLC, and LAS. This fine-grained evaluation framework allows us to analyze whether performance gains are consistent across different clinical dimensions or localized to specific error types.
Table~\ref{tab:subset_stats} summarizes the statistics of the VQA subsets.

\begin{table}[t]
\centering
\small
\setlength{\tabcolsep}{5pt}
\renewcommand{\arraystretch}{1.15}
\begin{tabular}{
  >{\raggedright\arraybackslash}p{0.13\columnwidth}@{\hspace{2pt}} llll
}
\toprule
\makecell[l]{\textbf{Error}\\\textbf{Type}} &
\multicolumn{1}{l}{\textbf{SLAKE}} &
\multicolumn{1}{l}{\textbf{VQA-RAD}} &
\multicolumn{1}{l}{\textbf{PathVQA}} &
\multicolumn{1}{l}{\textbf{Total}} \\
\midrule
MM  & 140 &  51 &  366 &  557 \\
SLC & 211 &  98 &  213 &  522 \\
AM  & 698 & 267 & 5268 & 6233 \\
LAS &  83 &  58 &  672 &  813 \\
\bottomrule
\end{tabular}
\caption{Screened question-answer pairs per error type and dataset.}
\label{tab:subset_stats}
\end{table}

\section{Hyperparameter Tuning}
SFT and DPO rely on different training objectives and, by design, their training data are not identical.
Specifically, SFT uses instruction-response pairs, whereas DPO uses preference pairs where the chosen $(m_w, y_w)$
matches the SFT data but requires additional curation of rejected responses. 
To enable a fair comparison, we therefore conducted independent hyperparameter searches using the SLAKE validation set to identify the best-performing
settings for each method (Table~\ref{tab:hp_search}).
Based on these analyses, we selected the following settings for our main experiments;
for SFT, we used a learning rate 2e-6 with 3 epochs, and for the DPO models, we used a learning rate 1e-7 with 3 epochs.

We further evaluated the effect of varying the number of training epochs (Table~\ref{tab:epoch_comp}),
confirming that performance gains are not attributable to additional training alone.

\begin{table}[t]
\centering
\small
\begin{tabular}{lccc}
\toprule
\textbf{lr} & \textbf{1 ep} & \textbf{2 ep} & \textbf{3 ep} \\
\midrule
\multicolumn{4}{l}{\textbf{\textit{SFT}}} \\
\midrule
1e-7  & 0.45 & 0.45 & 0.45 \\
5e-7  & 0.44 & 0.45 & 0.45 \\
1e-6  & 0.42 & 0.45 & 0.45 \\
2e-6  & 0.41 & 0.45 & 0.44 \\
2e-5  & 0.41 & 0.37 & 0.37 \\
\midrule
\multicolumn{4}{l}{\textbf{\textit{Text-Hallu + NLL}}} \\
\midrule
2e-8  & 0.44 & 0.45 & 0.45 \\
1e-7  & 0.46 & 0.45 & 0.47 \\
1e-6  & 0.45 & 0.47 & 0.47 \\
\bottomrule
\end{tabular}
\caption{Hyperparameter search results for LLaVA-Med on SLAKE validation set.}
\label{tab:hp_search}
\end{table}

\begin{table}[!t]
\centering
\footnotesize
\begin{tabular}{lccccc}
\toprule
\textbf{Model} & \textbf{1 ep} & \textbf{2 ep} & \textbf{3 ep} & \textbf{4 ep} & \textbf{5 ep} \\
\midrule
\multicolumn{6}{l}{\textbf{\textit{LLaVA-Med}}} \\
\midrule
Base Model                &   --  &   --  & 0.39 &   --  &   --  \\
SFT                 & 0.38 & 0.41 & 0.41 & 0.41 & 0.42 \\
\midrule
Text-Hallu         & 0.41 & 0.42 & 0.41 & 0.38 & 0.35 \\
\ \ + NLL           & 0.41 & 0.42 & 0.42 & 0.36 & 0.38 \\
Text-Noise          & 0.40 & 0.39 & 0.39 & 0.34 & 0.37 \\
\ \ + NLL           & 0.41 & 0.41 & 0.41 & 0.36 & 0.37 \\
IRPO                & 0.39 & 0.39 & 0.39 & 0.33 & 0.37 \\
\midrule
Image-Noise          & 0.40 & 0.40 & 0.40 & 0.36 & 0.37 \\
Image-ROI         & 0.41 & 0.41 & 0.41 & 0.34 & 0.38 \\
\midrule
mDPO                & 0.41 & 0.42 & 0.42 & 0.38 & 0.39 \\
MMedPO              & 0.40 & 0.39 & 0.40 & 0.32 & 0.33 \\
\midrule
\multicolumn{6}{l}{\textbf{\textit{HuatuoGPT-Vision}}} \\
\midrule
Base Model                &   --  &   --  & 0.49 &   --  &   --  \\
SFT                 & 0.51 & 0.50 & 0.52 & 0.51 & 0.52 \\
\midrule
Text-Hallu         & 0.52 & 0.51 & 0.53 & 0.51 & 0.49 \\
\ \ + NLL           & 0.52 & 0.51 & 0.52 & 0.49 & 0.50 \\
Text-Noise          & 0.52 & 0.50 & 0.51 & 0.53 & 0.51 \\
\ \ + NLL           & 0.50 & 0.52 & 0.52 & 0.52 & 0.51 \\
IRPO                & 0.51 & 0.50 & 0.53 & 0.53 & 0.51 \\
\midrule
Image-Noise          & 0.52 & 0.51 & 0.50 & 0.52 & 0.52 \\
Image-ROI         & 0.51 & 0.51 & 0.52 & 0.50 & 0.50 \\
\midrule
mDPO                & 0.52 & 0.51 & 0.51 & 0.54 & 0.53 \\
MMedPO              & 0.51 & 0.53 & 0.52 & 0.52 & 0.53 \\
\bottomrule
\end{tabular}
\caption{Performance comparison across epochs for LLaVA-Med and HuatuoGPT-Vision.}
\label{tab:epoch_comp}
\end{table}

\end{document}